# Complementary artificial intelligence designed to augment human discovery

Jamshid Sourati[a]
James Evans[a,b*]

[a]University of Chicago
1155 S. 60th Street
Chicago, IL 60637

[b]Santa Fe Institute
1399 Hyde Park Road
Santa Fe, NM 87501

Neither artificial intelligence designed to play Turing's imitation game, nor augmented intelligence built to maximize the human manipulation of information are tuned to accelerate innovation and improve humanity's collective advance against its greatest challenges. We reconceptualize and pilot beneficial AI to radically augment human understanding by complementing rather than competing with human cognitive capacity. Our approach to complementary intelligence builds on insights underlying the wisdom of crowds, which hinges on the independence and diversity of crowd members' information and approach. By programmatically incorporating information on the evolving distribution of scientific expertise from research papers, our approach follows the distribution of content in the literature while avoiding the scientific crowd and the hypotheses cognitively available to it. We use this approach to generate valuable predictions for what materials possess valuable energy-related properties (e.g., thermoelectricity), and what compounds possess valuable medical properties (e.g., asthma) that complement the human scientific crowd. We demonstrate that our complementary predictions, if identified by human scientists and inventors at all, are only discovered years further into the future. When we evaluate the promise of our predictions with first-principles or data-driven simulations of those properties, we demonstrate an "expectation gap" such that increased complementarity of our predictions do not decrease and in some cases increase the probability of these properties above those discovered and published by human scientists. In summary, by tuning AI to avoid the crowd, we can generate hypotheses unlikely to be imagined or pursued without intervention until the distant future that promise to punctuate scientific advance. By identifying and correcting for collective human bias, these models also suggest opportunities to improve human prediction by reformulating science education for discovery.

[*]Correspondence to jevans@uchicago.edu

Two competing visions for computational intelligence have dominated designs over the past half-century, but neither are tuned to accelerate humanity's advance against its greatest challenges, such as advancing science and technology for human benefit. Artificial Intelligence, coined by McCarthy in 1955, fixes humans as the standard of intelligence, following Turing's "imitation game"[1]. This influential approach became more tightly tethered to human intelligence with Samuel's work to build "machine learning" algorithms in the late 1950s[2] that not only produce human-like outcomes, but train on human moves. The Turing test vision of AI contrasted with a contemporary program to directly "Augment" Intelligence by reducing frictions in the conveyance and manipulation of information by Ashby[3], Englebart[4], Licklider[5] and others. If humanoid robots embody artificial intelligence; human-computer interfaces (e.g., screens, mice, EEG helmets, brain implants) realize augmented intelligence, but both visions feel inert to assist with science and technology design challenges, as in biomedicine and material science, on which millions of human scientists and engineers have collaborated and competed for centuries. Moreover, with millions of active scientists and engineers around the world and stagnating growth in labor productivity, halving to 1.3% for all but one OECD country since the 1990s[6], is the production of computational intelligence that mimics human capacity our most strategic or ethical investment? Here we reconceptualize and pilot beneficial AI that radically complements human understanding by thinking differently, complementing human cognitive capacity rather than competing with or directly extending it.

Our approach to complementary intelligence builds on insights underlying the wisdom of crowds[7], which hinges on the independence and diversity of crowd members' information[8] and approach[9]. In machine learning contexts like the Netflix Prize and Kaggle, ensemble models have always won[10]. In scientific crowds, findings established by more distinct methods and researchers are much more likely to replicate[11,12]. If we model discovery as establishing novel links among otherwise disconnected concepts[13], discovery cannot occur until discoverers arise with viewpoints that bridge the fields required to imagine those conceptual connections (Fig. 1a). This diversity of scientific viewpoints was implicitly drawn upon by pioneering information scientist Swanson in a heuristic approach to knowledge generation. For example, he hypothesized that if Raynaud's disorder was linked to blood viscosity in one literature, and fish oil was known to decrease that viscosity in another, then fish oil might lessen the symptoms of Raynaud's disorder, but would unlikely be arrived at within the field because no scientist was available to infer it[14–16], one of several hypotheses later experimentally demonstrated[17–19]. Expansive opportunities for discovery persist as researchers crowd around past discoveries[20], refusing to explore regions of knowledge cognitively distant from recent findings[21] (Extended Data Fig. 1). Our approach in this article scales and makes Swanson's heuristic continuous, combining it with explicit measurement of the scientific expertise distribution that draw upon advances in unsupervised manifold learning[22]. Recent efforts to generate scientific hypotheses rely heavily on scientific literature, but ignore equally available publication meta-data. By programmatically incorporating information on the evolving distribution of scientific expertise, our approach targets the exploration of areas far from past discoveries, avoiding the scientific crowd. As such, the suggestions that result complement collective intelligence and enable us to punctuate advance by identifying promising experiments unlikely to be pursued by scientists in the near future without intervention.

In order to avoid the scientific crowd, our approach must first identify those topics at the focus of collective attention in the scientific system. Metadata about the distribution of research experts across topics and time represents a critical social fact that can stably improve our inference about whether scientific relationships will receive scientific attention or remain unimagined and unexplored[13,23]. We build expert awareness into our approach to identify and validate the scientific and technological benefit of pursuing complementary research avenues unlikely to be considered by unassisted human experts. The proposed framework provides opportunities for intellectual arbitrage between isolated communities through complementary intelligences unconstrained by the human incentive to flock together within fields.

**Avoiding Cognitive Availability**

We model the cognitive availability of a hypothesis to human scientists by measuring the distribution of experts exposed to its underlying concepts, linked by previous discoveries that intermediate them and which could guide human intuition from one to the other (Fig. 1a). The distribution of relevant experts in science can be estimated from a sufficient corpus of research articles, where papers inscribe the mixed network of publishing scientists and concepts they investigate. We represent these complex connectivities with a hypergraph, where published articles are hyperedges connecting authors and mentioned concepts. Hypergraphs are effective at representing complex social interaction[24–26] and proximity between concepts across them quantifies their cognitive availability to scientist teams, which effectively forecasts human discovery and publication[27]. Scientific entities further apart in the hypergraph will be less likely conceived together, or seen as relevant by scientists, dramatically reducing their chance for consideration and discovery. We can measure node proximities with any graph distance metric that varies with expert density, such as unsupervised neural embeddings, Markov transition probabilities, or self-avoiding walks from Schramm-Loewner evolutions. Here we use shortest-path distances (SPD) between conceptual nodes, as interlinked by authors in our mixed hypergraph. In the remainder of our paper, we divide concepts into materials such as chemical compounds and the valuable scientific properties that may be attributed them, like conductivity, treatment potential, regulation of a disease-related gene, etc. The hypotheses we explore involve material and biomedical relationships between materials and their properties.

In order to avoid selecting hypotheses without scientific promise, cognitive availability must couple with a signal of hypothesis plausibility. Such a signal could be provided by the published research literature and quantified with unsupervised knowledge embedding models[28]. Alternatively, a signal of plausibility could be derived from theory-driven models of material properties. Here we use unsupervised knowledge embeddings for our algorithm, reserving theory-driven model simulations to evaluate the value and human complementarity of our predictions. Specifically, we measure the scientific merit of any given hypothesis using the cosine similarity between embedding vectors of material and property nodes that comprise each hypothesis. Figure 1b provides a general overview of our algorithm for inferring materials with a target property. Initialized once a pool of candidate materials has been extracted from literature, we perform parallel operations to generate hypotheses that are both scientifically plausible and human-complementary. We train an unsupervised word embedding model over prior publications and measure scientific relevance as cosine distance in the embedding. In parallel, we indicate cognitive availability by structuring the hypergraph such that each author and material or property node from a paper is encased within a hyperedge and shortest path distances between the property and all materials are computed across the graph. We transform signals of plausibility and cognitive availability into a unified scale and linearly combine them with a mixing coefficient $\beta$ (see details in Methods and Supplementary Information). With its expert awareness, our algorithm can symmetrically generate either the most or least-human hypotheses—those likely to compete versus complement collective human capacity—based on the sign of the mixing coefficient. Negative $\beta$ values lead to predictions that mimic human experts in discovery, while positive values produce hypotheses least similar to those human experts could infer, straddling socially but not scientifically disconnected fields. At extremes, $\beta$=-1 and 1 yield algorithms that generate predictions very familiar or very alien to human experts, regardless of scientific merit. Setting $\beta$=0 implies exclusive emphasis on scientific plausibility, blind to the distribution of experts. This mode is equivalent to traditional discovery prediction methods exclusively based on previously published content. Intermediate positive $\beta$s balance exploitation of relevant materials with exploration of areas unlikely considered or connected by human experts. Materials with the highest resulting scores are reported as the algorithm's predictions.

In the following sections, we evaluate the complementarity of our inferences for human science by verifying (1) their distinctness from contemporary investigations and (2) their scientific promise. We anticipate that both features will simultaneously increase in ranges of $\beta$ higher than those that characterize published science. Scientific merit will naturally reduce at the extremes of our interval [-1,1], however, where the algorithm ignores an inferred hypothesis' literature-based plausibility.

**Evaluating Discovered Predictions**

As we increase $\beta$, the algorithm avoids inferences that lie within regions of high expert density and focuses on candidate materials and properties that span disciplinary divides and evade human attention. As a result, we expect that generated hypotheses with large $\beta$ will diverge from those pursued by the scientific community, will less likely become published, and if published, will be discovered further into the future, after science has reorganized itself to consider them. In order to verify these hypotheses, we first assess the discoverability of hypotheses inferred from different $\beta$ values by computing the precision between our inferences and published discoveries. Results strongly confirm our expectation that materials inferred at higher $\beta$ values are less discoverable by human scientists (Extended Data Fig. 2). Materials distant from a given property in the hypergraph remain cognitively unavailable to scientists in the property's proximity (Fig. 1c). It takes longer for researchers in the field to broach knowledge gaps separating unfamiliar materials from valued properties. Among the inferences eventually discovered, we measure the discovery waiting time and expect to observe an increasing trend in wait times as we move from negative (human-competitive) to positive (human-complementary) $\beta$ values in our predictions. Generating 50 hypotheses per $\beta$ value and evaluating the resulting predictions indicates that for the majority of targeted properties the average discovery wait times climb markedly when increasing $\beta$ (Fig. 2) for energy-related chemical properties (Fig. 2a-2c), COVID-19 (Fig. 2d) and 70% of the other human diseases (Fig. 2e). Averaging wait times across all human diseases manifests a clear increasing trend. For some cases such as COVID-19 (Fig. 2d), none of the complementary predictions made with positive $\beta$ values come to be discovered by humans within the time frame we examine.

**Evaluating Undiscovered Predictions**

To evaluate the scientific merit of our algorithm's undiscovered hypotheses requires data beyond the extant literature. Such hypotheses necessarily grow to comprise the vast majority of cases for large values of $\beta$. If science was an efficient market and experts optimally pursued scientific quality, then in human-avoiding high $\beta$ hypotheses, we would observe a proportional decline in their scientific promise and efficacy. On the other hand, if scientists crowd together along the frontier of scientific possibility and their continued efforts yield diminishing marginal returns, we might observe an increase in promise as we move beyond them.

To evaluate the merit of scientific inferences, we utilize first-principles or data-driven models derived uniquely for each property based on well-established theoretical principles within the field. Such models assign real-valued scores to candidate materials as a measure of their potential for possessing the targeted properties. These computations may be carried out without regard for whether materials have yet been discovered, making them a suitable, if conservative, scoring function for evaluating undiscovered hypotheses. We produced such scores for approximately 45% of the properties we considered above using first-principle equations or based on databases curated through high-throughput protein screens. To evaluate thermoelectric promise, we used power factor (PF) as an important component of the overall thermoelectric figure of merit, $zT$, calculated using density functional theory for candidate materials as a strong indication of thermoelectricity[29,30]. To evaluate ferroelectricity, estimates of spontaneous polarization obtained through symmetry analysis and relevant theoretical equations serve as a reliable metric for this property[31]. For human diseases including COVID-19, proximity between disease agents (e.g., SARS-CoV-2) and candidate compounds in protein-protein interaction networks suggests the likelihood a material will recognize and engage with the disease agent[32] (for more details on how these theoretical scores are derived see the Supplementary Information). We note that scores based on first-principles equations or simulations represent conservative estimates of scientific merit as they are based on widely-accepted, scientist-crafted and theory-inspired models. Because these scores are potentially available to scientists in the area, they may be considered when guiding investigation, such that experiments on these unevaluated hypotheses are very often promising. Nevertheless, in what follows we show that intermediate positive $\beta$s manifest continuation or improvement on even this conservative measure of quality.

We expect the average theoretical scores of hypotheses to decay significantly at the extremes of the $\beta$ range [-1,1], as at those points the algorithm ignores the merit signal putting it at higher risk of generating scientifically irrelevant (or absurd) proposals. We expect, however, that this decay will occur more slowly than the decrease in hypothesis discovery and publication, which implies a $\beta$ interval where proposals are not discoverable but highly

promising—an ideal operating region for the generation of hypotheses that complement those from the human scientific crowd. In order to verify this, we contrasted changes in average theoretical scores with the discoverability of generated hypotheses for various $β$ values, which we quantify with precision—the overlap between predictions and published discoveries. As illustrated in Fig. 3 (first row), discoverability decreases near the transition of $β$ from negative to positive values, but its decay is much sharper than average theoretical scores, which do not collapse until nearly $β=0.4$. This holds for electrochemical properties and the majority of diseases. Results for certain individual diseases can be seen in the second row of Fig. 3 (for the full set of results see Extended Data Fig. 3 and Supplementary Information). Moreover, note that for the cases investigated, average theoretical scores for inferred hypotheses grow higher than average scores for actual, published discoveries before eventual decay at high $β$ values. For certain properties like thermoelectricity or therapeutic efficacy against the disease Alopecia, theoretical merit of our inferences exhibit striking and dramatic growth from negative (scientist-mimicking) to positive (scientist-avoiding) hypotheses, suggesting strong diminishing returns to following these scientific crowds, whose overharvested fields have become barren for new discoveries.

In order to compare the decay rate of discoverability and theoretical scores, we define and compute the *expectation gap* to measure the distance between expected values for two conditional distributions over $β$. A randomly selected prediction is (1) identified as promising based on its corresponding first-principle score, and (2) discoverable, i.e., studied and published by a scientist following prediction year (for details see Methods and Supplementary Information). A positive expectation gap indicates that increasing $β$ will preserve the quality of predictions while making them more complementary to human hypotheses. As shown in Fig. 4a, the vast majority of properties considered in this section yield substantial and significantly positive expectation gaps. Building on this, we use a probabilistic model to assess the complementarity of our algorithm's prediction with those of the scientific community for any value of $β$. This is done by computing the joint probability that a randomly selected prediction is plausible in terms of the desired property and beyond current scientists' scope of research (see Supplementary Information). These probabilities specify the optimal $β$ to balance exploitation and exploration in augmenting collective human prediction. Results in Fig. 4b indicates the optimal point varies for different properties, but one can distinguish the range 0.2-0.3 as the most consistently promising interval.

**Discussion**

Here we explore the potential for building AI algorithms to radically augment the scientific community. Building on insights about independence underlying the wisdom of crowds, we seek to complement the clustering driven by interactions and institutions of the scientific community. By tuning our algorithm to avoid the crowd, we generate promising hypotheses unlikely to be imagined, pursued or published without machine recommendation for years into the future. By identifying and correcting for collective patterns of human attention, formed by field boundaries and institutionalized education, these models complement the contemporary scientific community. A further class of complementary predictions could be tuned to compensate not only for emergent collective bias, but universal cognitive constraints, such as limits on the human capacity to conceive or search through complex combinations (e.g., high-order therapeutic cocktails[33]). Disorienting hypotheses from such a system will not be beautiful, but being inconceivable, they break fresh ground and sidestep the path-dependent "burden of knowledge" where scientific institutions require new advances built upon the old for ratification and support[34,35].

Our approach can also be used to identify individual and collective biases that limit productive exploration and suggest opportunities to improve human prediction by reformulating science education for discovery. Insofar as research experiences and relationships condition the questions scientists investigate, education tuned to discovery would conceive of each student as a new experiment, recombining knowledge and opportunity in novel ways. Our investigation underscores the power of incorporating human and social factors to produce artificial intelligence that complements rather than substitutes for human expertise. By making AI hypothesis generation aware of human expertise, it can race with rather than against the scientific community to expand the scope of human imagination and discovery.

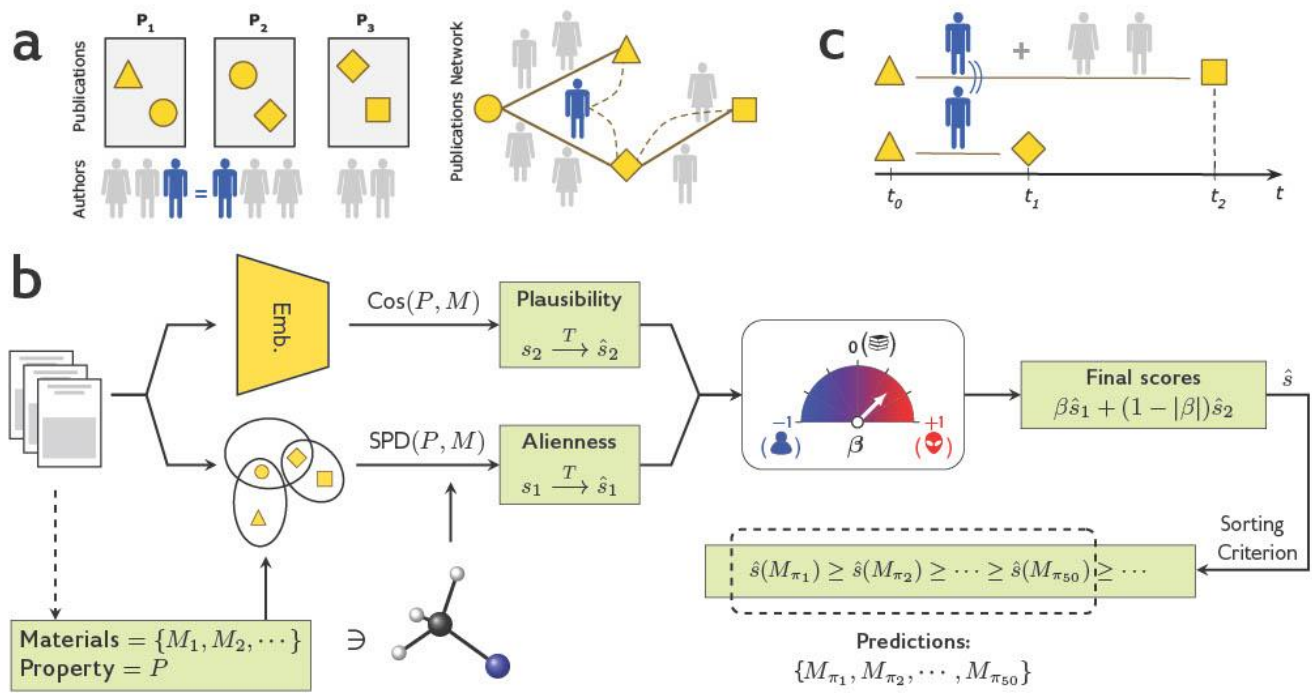

**Fig. 1. (a)** Distribution and overlap of experts investigating (and publishing on) topics represented by yellow geometric shapes. Dashed lines represent paths of more or less cognitive availability between topics ("triangle", "diamond" and "square"). **(b)** Overview of our complementary discovery prediction algorithm. Beginning with a scientific corpus and a targeted property, candidate materials are extracted from the corpus and used along with property mentions and authors to form the hypergraph. The algorithm follows two branches to compute plausibility from word embedding semantic similarities and "alienness" or human inaccessibility from hypergraph shortest-path distances. These two signals are combined after proper normalization and standardization through the mixing coefficient $\beta$ to generate a prediction more or less complementary to the flow of human discovery. Candidate materials are sorted based on resulting scores and those with highest rank are reported as proposed discoveries. **(c)** Discovery wait times for relations between "triangle"–"diamond" and "triangle"–"square". The time one needs to wait for a relationship to be discovered is proportional to the path length of cognitive availability between the two relevant topics. The denser presence of experts around the pair "triangle"–"diamond" implies greater cognitive availability leading to earlier discovery and publication versus "triangle"–"square" where the connection requires a longer path.

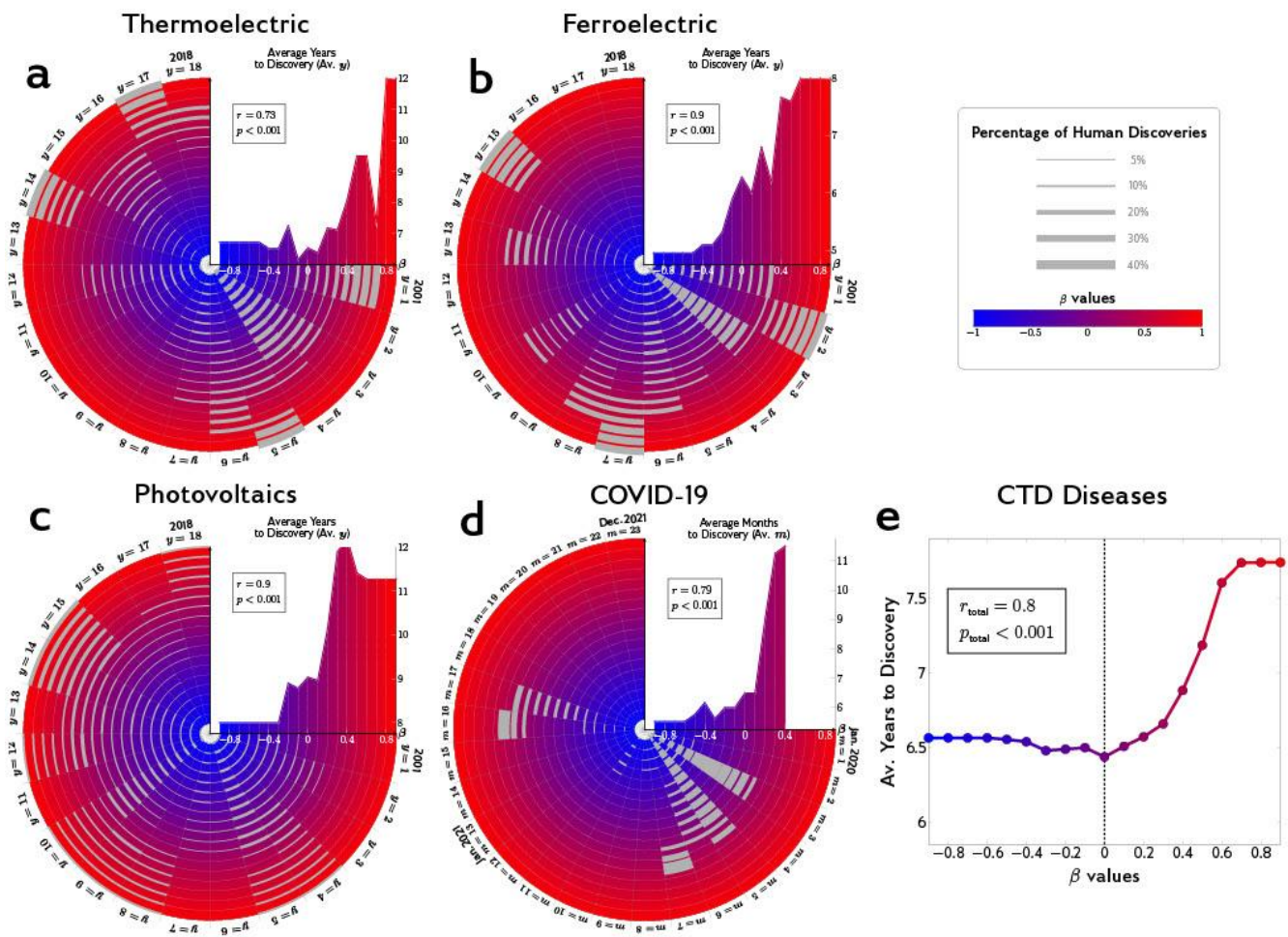

**Fig. 2.** Wait time for published discoveries associated with distinct properties and different $\beta$ values. **(a-d)** Average annual/monthly discovery wait times are shown as thick gray arcs, where thickness represents the percentage of materials discovered in the corresponding year/month. Each orbit is associated with a particular $\beta$ value with larger (more red) orbits representing larger $\beta$ values. The values we consider here vary between -0.8 (the smallest, bluest orbit) and 0.8 (the largest, reddest orbit). The plot in the upper right quarter of the orbits reveals the total average of discovery wait times including all years/months for the considered $\beta$ values. **(f)** Total average for wait times across all the human diseases (except COVID-19) in our experiments.

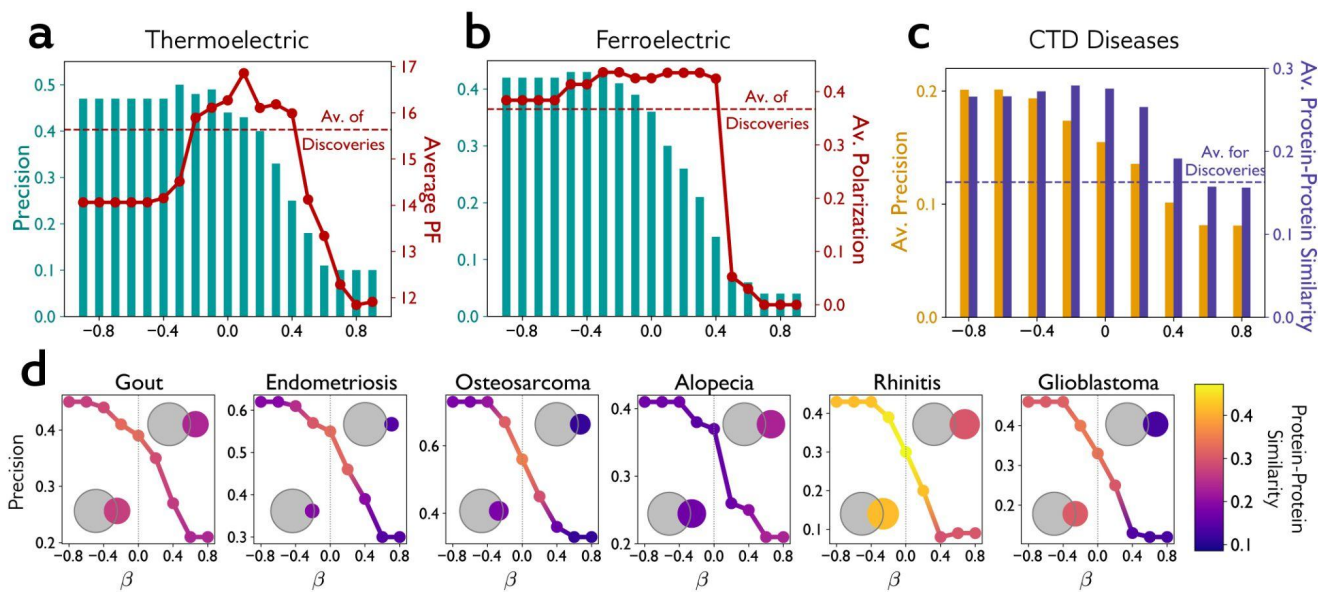

**Fig. 3.** Overlapping percentage and average theoretical scores calculated for predictions. **(a-b)** Green bars show overlapping percentages and curves indicate (a) average PF for thermoelectricity and (b) spontaneous polarization for ferroelectricity. **(c)** Overlapping and average theoretical scores (i.e., protein-protein interaction similarity scores) of the therapeutic predictions. Dashed lines in all cases show average theoretical scores computed for actual discoveries following prediction year. **(d)** Overlapping versus average protein-protein similarity scores for nine human disease examples. The *y*-axis indicates overlapping percentage and color gradient represents average theoretical scores for predictions.

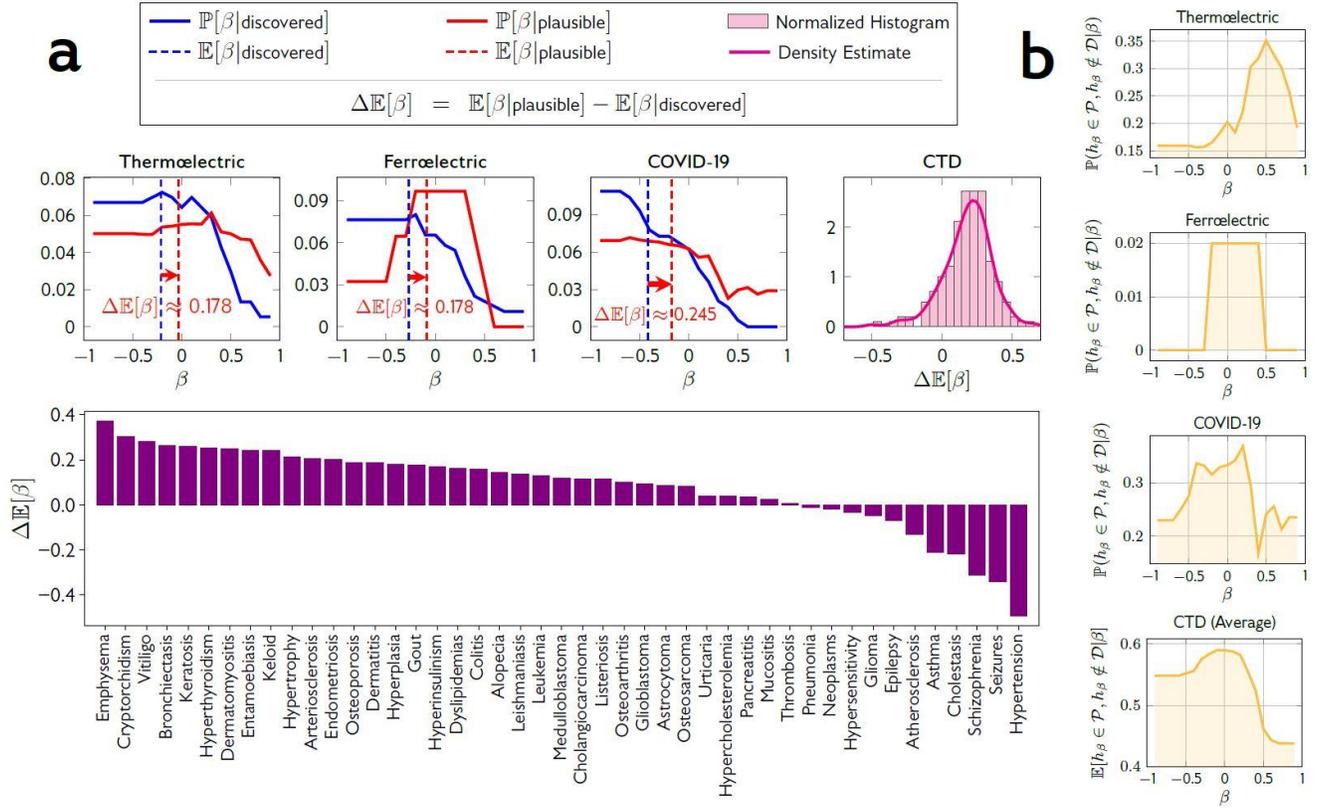

**Fig. 4. (a)** Expectation gap calculated for properties with theoretical first-principle scores. We plot the conditional distributions $\mathbb{P}(\beta|\text{plausible})$ and $\mathbb{P}(\beta|\text{discoverable})$ separately for Thermoelectricity, Ferroelectricity and COVID-19, whereas for the remainder of human diseases included in our experiments we simply show the normalized histogram (first row) and individual (second row) gaps. **(b)** The joint probability of simultaneous undiscoverability and plausibility for different $\beta$ values.

# Methods

## Experiments and Data Collection

We used two distinct datasets in our experiments. For energy-related properties, i.e., thermoelectricity, ferroelectricity and photovoltaic materials, we used a pre-curated dataset of approximately 1.5M articles whose topics are relevant to inorganic materials. These articles have been selected and pre-processed by Tshitoyan et. al (2019)[28], who also made their DOIs publicly available. We downloaded abstracts of these DOIs through the Scopus API provided by Elsevier (https://dev.elsevier.com/) and extracted 106K candidate inorganic materials from the downloaded abstracts using Python Materials Genomics[36] and direct rule-based string processing. For COVID-19 and other human diseases, we used the MEDLINE database which includes more than 28M articles published on a wide range of topics. In this dataset, we identified around 7,800 approved candidate drugs, from which we selected approximately 4,000 drugs with simple names (excluding names with multiple numerical subparts). We use Comparative Toxicogenomics Database (CTD)[37] to extract ground-truth associations between our drug pool and 400 human diseases (besides COVID-19), selected such that they represent the largest number of associations. Note that in order to form our hypergraph, we need to know who authored the articles. The Scopus API distinguishes distinct authors and assigns unique codes to them. However, this is not the case with MEDLINE, where authors are not identified other than by name. We use the set of disambiguated authors shared through PubMed Knowledge Graph (PKG) package[38], which were obtained by combining results from the Author-ity disambiguation of PubMed[39] and the more recent semantic scholar database[40].

Our discovery prediction experiment begins by setting a date of prediction (e.g., the beginning of January 2001). We then form our hypergraph using literature prior to that date and let our algorithm make predictions from materials unstudied in relation to a given property at that point. Many of our evaluation criteria are based on human discovery. For energy-related properties, we model human discovery as first-time co-occurrence of materials with the targeted property, following methodology of the team that curated the dataset[28]. For all diseases except COVID-19, human discoveries were identified through drug-disease associations indicated in CTD. We set the date for each drug-disease discovery to the earliest publication reported by the CTD for curated associations. For COVID-19, discovered drugs are identified based on their involvement in COVID-related studies reported by ClinicalTrials.org that began after breakout of the disease in the US in the beginning of 2020. Discovery date for each association is set to the date the corresponding study was first posted, and if the drug was involved in multiple trials we considered the earliest. There were 6,280 trials posted as of August 5th, 2021 (ignoring 37 trials dated before 2020), which included 279 drugs from our pool (~7%) within their designs.

## Prediction Algorithm

Our predictor consists of two scoring functions. The first measures the cognitive unavailability ("alienness") of candidate materials via Shortest-Path distance (SPD) between the nodes corresponding to the targeted property and candidates. The second measures scientific plausibility through the semantic cosine similarities of their corresponding keywords. For this purpose, we train skipgram word2vec embedding models over the literature (literature collected on inorganic materials for energy-related properties and MEDLINE for the diseases) produced prior to the prediction year. The prediction year is set to the beginning of 2001 for all the considered properties except for COVID-19 for which the prediction year is set to the beginning of 2020. We combine the alienness and plausability scores with a mixing coefficient, denoted by $β$, adjusting their contributions to obtain a final score for the candidate. The plausibility component yields continuous scores distributed close to Gaussian, whereas the alienness component offers unbounded ordinal SPD values. Simple normalization methods are insufficient to combine scores with such distinct characteristics. As a result, we first standardize the two scores to a unified scale by applying the Van der Waerden transformation[41], followed by a Z-score normalization. The final step includes taking the weighted average of the resulting Z-scores with weights depending on *β* (see Supplementary Information for more details).

We want our predictor to infer undiscoverable yet promising hypotheses. Setting *β* to a more positive value makes predictions less familiar and more alien, i.e., less discoverable. Moreover, increasing *β* to the positive extreme (i.e., +1) excludes scientific merit from the algorithm's objective in materials selection. Hence, growing *β* causes both discoverability and plausibility of predictions to decay. What matters to us is that plausibility decreases more slowly than discoverability, suggesting that the predictor achieves a close-to-ideal state where predictions are simultaneously alien and promising. In order to verify this with a single number, we define the *expectation gap* criterion, computed as the difference between expected values of the following two distributions over *β*: $\mathbb{P}(\beta|\text{plausible})$ and $\mathbb{P}(\beta|\text{discoverable})$. The terms "plausible" and "discoverable" on the conditional sides could be substituted by the precise statements "a randomly selected inferred hypothesis is theoretically plausible" and "a randomly selected inferred hypothesis is discoverable"—it will be published by scientists, respectively. While we know both of these distributions reduce as *β* approaches +1, the expectation gap measures any positive shift in the mass of $\mathbb{P}(\beta|\text{plausible})$ against $\mathbb{P}(\beta|\text{discoverable})$. The likelihood of discovery $\mathbb{P}(\beta|\text{discoverable})$ can be estimated through an empirical distribution of predictions discovered and published. Scientific plausibility can be estimated by leveraging properties' theoretical scores obtained from prior knowledge and first-principles equations and data from relevant fields. We estimate $\mathbb{P}(\beta=\beta_0|\text{ plausible})$ in two steps: (1) converting theoretical scores to probabilities, and (2) computing weighted maximum likelihood estimates of $\mathbb{P}(\beta=\beta_0|\text{plausible})$ given a set of predictions generated by our algorithm operated with $\beta_0$ (see Supplementary Information for details). We restrict experiments in this section to only those properties for which we could obtain a reliable source of theoretical scores (see Supplementary Information for details of the scores): thermoelectricity, ferroelectricity, COVID-19 and 175 other human diseases (178 out of 404 total properties). Finally, note that expectation gaps and average discovery dates (described above) say nothing about the *β* interval most likely to lead to complementarity and plausibility. We introduce an additional probabilistic criterion for this purpose, which jointly models these two features and computes their likelihood for various *β* values, $\mathbb{P}(\text{undiscoverable, plausible} | \beta)$. One can use this distribution to screen the best operating point for complementary artificial intelligence (see Supplementary Information).

# References

1. Turing, A. M. I.—COMPUTING MACHINERY AND INTELLIGENCE. *Mind* vol. LIX 433–460 (1950).

2. Samuel, A. L. Some Studies in Machine Learning Using the Game of Checkers. *IBM Journal of Research and Development* vol. 3 210–229 (1959).

3. Ashby, W. R. An introduction to cybernetics. (1956) doi:10.5962/bhl.title.5851.

4. Engelbart, D. C. Augmenting human intellect: A conceptual framework. *Menlo Park, CA* (1962).

5. Licklider, J. C. R. Man-Computer Symbiosis. *IRE Transactions on Human Factors in Electronics* **HFE-1**, 4–11 (1960).

6. Brynjolfsson, E., Rock, D. & Syverson, C. *Artificial Intelligence and the Modern Productivity Paradox: A Clash of Expectations and Statistics*. https://www.nber.org/papers/w24001 (2017) doi:10.3386/w24001.

7. Galton, F. Vox populi (the wisdom of crowds). *Nature* **75**, 450–451 (1907).

8. Surowiecki, J. The wisdom of crowds: Why the many are smarter than the few and how collective wisdom shapes business. *Economies, Societies and Nations* **296**, (2004).

9. Page, S. E. *The Diversity Bonus: How Great Teams Pay Off in the Knowledge Economy*. (Princeton University Press, 2019).

10. Freund, Y., Schapire, R. E. & Others. Experiments with a new boosting algorithm. in *icml* vol. 96 148–156 (Citeseer, 1996).

11. Danchev, V., Rzhetsky, A. & Evans, J. A. Centralized scientific communities are less likely to generate replicable results. *Elife* **8**, (2019).

12. Belikov, A. V., Rzhetsky, A. & Evans, J. Prediction of robust scientific facts from literature. *Nature Machine Intelligence* 1–10 (2022).

13. Rzhetsky, A., Foster, J. G., Foster, I. T. & Evans, J. A. Choosing experiments to accelerate collective discovery. *Proc. Natl. Acad. Sci. U. S. A.* **112**, 14569–14574 (2015).

14. Swanson, D. R. Fish oil, Raynaud's syndrome, and undiscovered public knowledge. *Perspect. Biol. Med.* **30**, 7–18 (1986).

15. Swanson, D. R. Medical literature as a potential source of new knowledge. *Bull. Med. Libr. Assoc.* **78**, 29–37 (1990).

16. Weeber, M., Klein, H., de Jong-van den Berg, L. T. W. & Vos, R. Using concepts in literature-based discovery: Simulating Swanson's Raynaud--fish oil and migraine--magnesium discoveries. *J. Am. Soc. Inf. Sci. Technol.* **52**, 548–557 (2001).

17. Evans, J. & Rzhetsky, A. Machine Science. *Science* **329**, 399–400 (2010).

18. Digiacomo, R. A., Kremer, J. M. & Shah, D. M. Fish-oil dietary supplementation in patients with Raynaud's phenomenon: A double-blind, controlled, prospective study. *The American Journal of Medicine* vol. 86 158–164 (1989).

19. Chiu, H.-Y., Yeh, T.-H., Huang, Y.-C. & Chen, P.-Y. Effects of Intravenous and Oral Magnesium on Reducing Migraine: A Meta-analysis of Randomized Controlled Trials. *Pain Physician* **19**, E97–112 (2016).

20. Singer, U., Radinsky, K. & Horvitz, E. On Biases Of Attention In Scientific Discovery. *Bioinformatics* (2020)

doi:10.1093/bioinformatics/btaa1036.

21. Chu, J. S. G. & Evans, J. A. Slowed canonical progress in large fields of science. *Proc. Natl. Acad. Sci. U. S. A.* **118**, (2021).

22. Mikolov, T., Yih, W. & Zweig, G. Linguistic regularities in continuous space word representations. *hlt-Naacl* (2013).

23. Sourati, J., Belikov, A. & Evans, J. Data on how science is made can make science better. (2022).

24. Shi, F., Foster, J. G. & Evans, J. A. Weaving the fabric of science: Dynamic network models of science's unfolding structure. *Soc. Networks* **43**, 73–85 (2015).

25. Lung, R. I., Gaskó, N. & Suciu, M. A. A hypergraph model for representing scientific output. *Scientometrics* **117**, 1361–1379 (2018).

26. Antelmi, A., Cordasco, G., Spagnuolo, C. & Szufel, P. Social Influence Maximization in Hypergraphs. *Entropy* **23**, (2021).

27. Sourati, J. & Evans, J. Accelerating science with human versus alien artificial intelligences. *arXiv [cs.AI]* (2021).

28. Tshitoyan, V. *et al.* Unsupervised word embeddings capture latent knowledge from materials science literature. *Nature* **571**, 95–98 (2019).

29. Mehdizadeh Dehkordi, A., Zebarjadi, M., He, J. & Tritt, T. M. Thermoelectric power factor: Enhancement mechanisms and strategies for higher performance thermoelectric materials. *Mater. Sci. Eng. R Rep.* **97**, 1–22 (2015).

30. Ricci, F. *et al.* An ab initio electronic transport database for inorganic materials. *Sci Data* **4**, 170085 (2017).

31. Smidt, T. E., Mack, S. A., Reyes-Lillo, S. E., Jain, A. & Neaton, J. B. An automatically curated first-principles database of ferroelectrics. *Sci Data* **7**, 72 (2020).

32. Gysi, D. M. *et al.* Network Medicine Framework for Identifying Drug Repurposing Opportunities for COVID-19. *ArXiv* (2020).

33. Gediya, L. K. & Njar, V. C. Promise and challenges in drug discovery and development of hybrid anticancer drugs. *Expert Opin. Drug Discov.* **4**, 1099–1111 (2009).

34. Jones, B. F. The Burden of Knowledge and the 'Death of the Renaissance Man': Is Innovation Getting Harder? *Rev. Econ. Stud.* **76**, 283–317 (2009).

35. Szell, M., Ma, Y. & Sinatra, R. A Nobel opportunity for interdisciplinarity. *Nat. Phys.* **14**, 1075–1078 (2018).

36. Ong, S. P. *et al.* Python Materials Genomics (pymatgen): A robust, open-source python library for materials analysis. *Comput. Mater. Sci.* **68**, 314–319 (2013).

37. Davis, A. P. *et al.* The Comparative Toxicogenomics Database: update 2019. *Nucleic Acids Res.* **47**, D948–D954 (2019).

38. Xu, J. *et al.* Building a PubMed knowledge graph. *Sci Data* **7**, 205 (2020).

39. Torvik, V. I. & Smalheiser, N. R. Author Name Disambiguation in MEDLINE. *ACM Trans. Knowl. Discov. Data* **3**, (2009).

40. Ammar, W. *et al.* Construction of the Literature Graph in Semantic Scholar. *arXiv [cs.CL]* (2018).

41. Coakley, C. W. Practical Nonparametric Statistics (3rd ed.). *J. Am. Stat. Assoc.* **95**, 332 (2000).

# Acknowledgements

The authors wish to thank our funders for their generous support: National Science Foundation #1829366; Air Force Office of Scientific Research #FA9550-19-1-0354, #FA9550-15-1-0162; DARPA #HR00111820006. We thank Laszlo Barabasi and Deisy Morselli Gysi for helpful data related to their network-based forecast of COVID-19 drugs and vaccines with protein-protein interactions [32], and Anubhav Jain, Vahe Tshitoyan and Alex Dunn for sharing data and code to help replicate their work on unsupervised word embeddings and latent knowledge about material science[28]. We also thank participants of the Santa Fe Institute workshop "Foundations of Intelligence in Natural and Artificial Systems", the University of Wisconsin at Madison's HAMLET workshop, and colleagues at the Knowledge Lab for helpful comments.

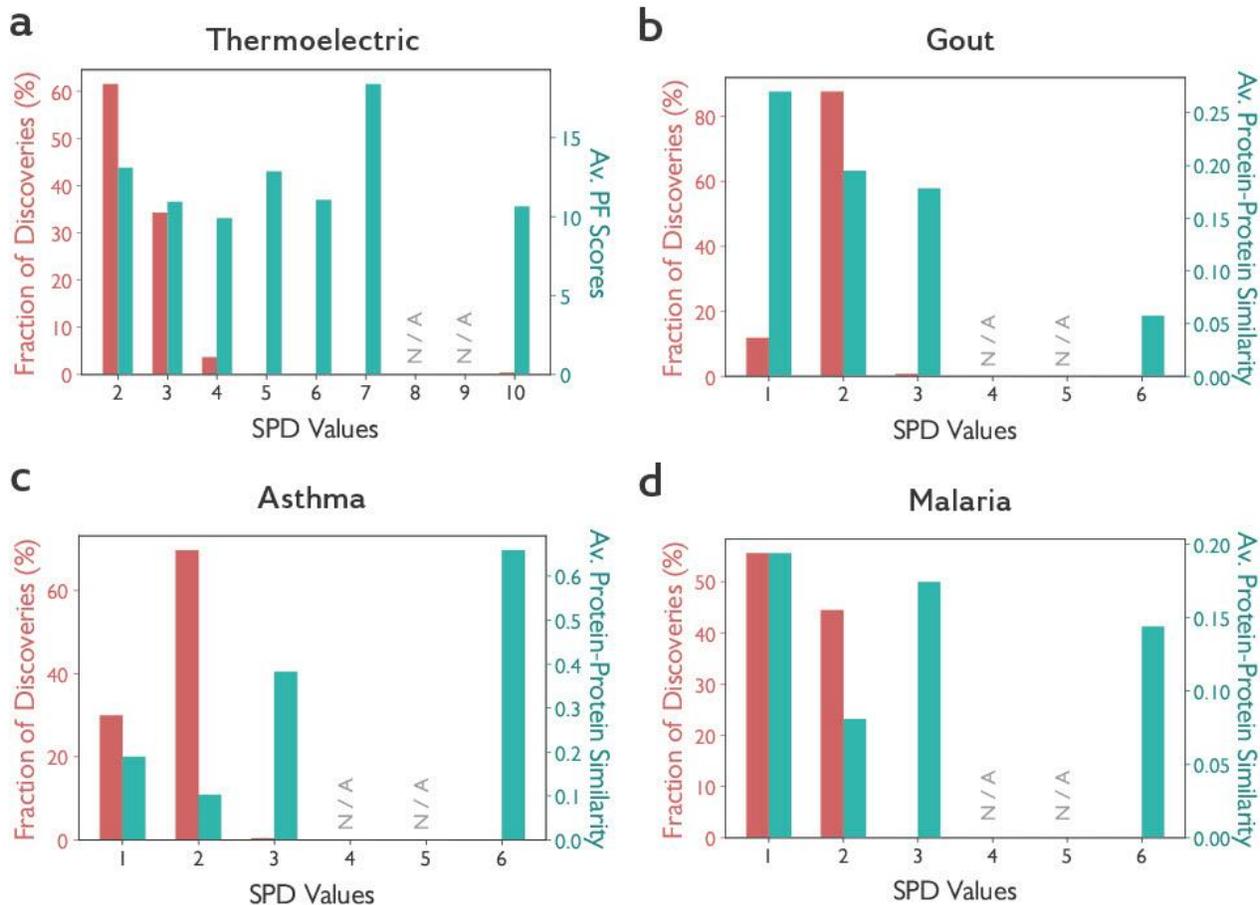

**Extended Data Fig. 1.** Illustration of localized discoveries made by scientists regarding thermoelectric materials (a) and repurposing materials for treating gout (b), asthma (c) and malaria (d). Red bars indicate fractions of discoveries occurring at various levels of proximity (measured through shortest path distances (SPD) in a literature-based hypergraph) to a particular targeted property. Note how these distributions concentrate around low proximites. Blue bars indicate average scores representing plausibility that candidate materials have the targeted property in theory. For thermoelectricity (a), we defined Power Factor (PF) as the plausibility score, and for the three human diseases shown here (b-d), scores are obtained through similarities between protein profiles of the candidate materials and the targeted diseases.

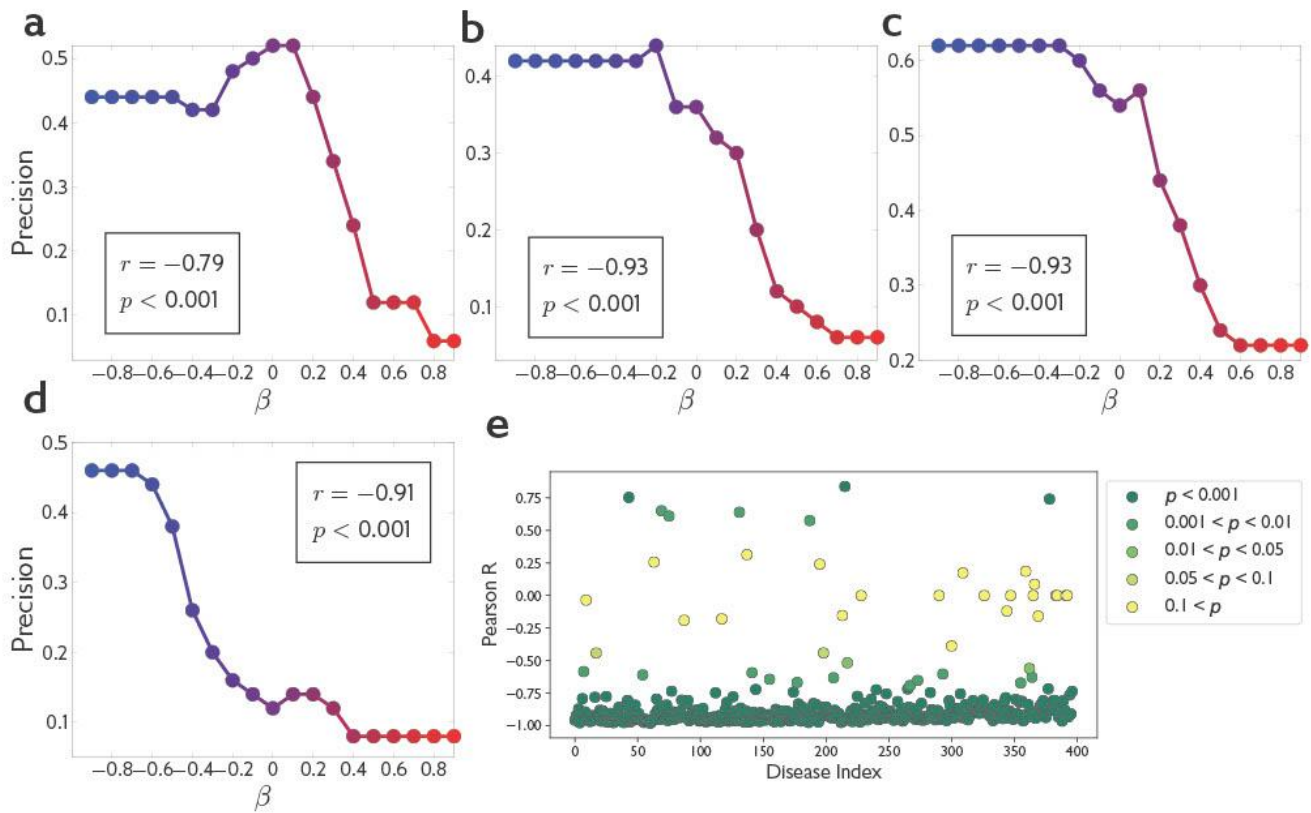

**Extended Data Fig. 2.** Illustration of decaying discoverability for predictions as $\beta$ increases. Discoverability of predictions is measured through computing the precision metric, i.e., their overlapping percentage with respect to actual discoveries made after prediction year. Decreasing precision curves and their highly negative Pearson correlation coefficients are shown for (a) thermoelectricity, (b) ferroelectricity, (c) photovoltaics and (d) COVID-19. We visualize these statistics for the remaining human diseases with a scatterplot of their Pearson correlation coefficients (e).

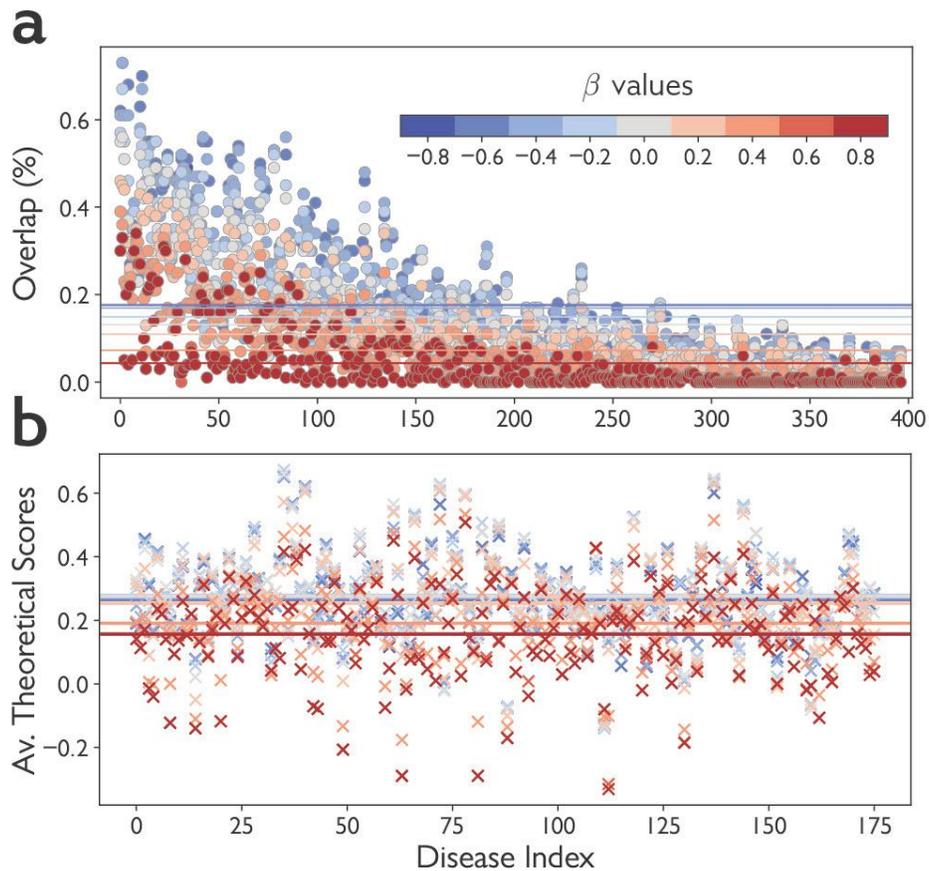

**Extended Data Fig. 3.** Discoverability and scientific merit for predictions made with varying $\beta$ values in the research case repurposing drugs for treating human diseases. (a) Precision values for predictions generated with eight levels of $\beta$ and computed for all 400 human diseases we considered (except COVID-19). Diseases are sorted in terms of the number of relevant drugs. (b) Average theoretical scores measured through protein-protein similarity between diseases and candidate drugs for predictions generated with the same $\beta$ values. We compute such protein-based theoretical scores for 176 diseases out of 400 total cases (44%). In both subfigures, horizontal lines show average values across all diseases.

# SUPPLEMENTARY INFORMATION: COMPLEMENTARY ARTIFICIAL INTELLIGENCE DESIGNED TO AUGMENT HUMAN DISCOVERY

Jamshid Sourati[1], James A. Evans[1,2]

[1]Knowledge Lab, University of Chicago, Chicago, IL, USA

[2]Santa Fe Institute, Santa Fe, NM, USA

## S1   Combining Scores

Our algorithm combines two sources of information for measuring human availability and scientific plausibility, each of which separately scores the candidate materials. The two scores will then be combined through $\beta$ to result in a single scalar score such that the magnitude of $\beta$ determines the weight of whether human availability is encouraged (negative) or discouraged (positive) among predictions, as opposed to scientific plausibility (0). As $\beta$ grows from one extreme ($-1$) to the other ($+1$) the algorithm lessens its objective of human accessibility, instead avoiding it after passing $\beta = 0$. Hence, at least three special operating points are distinguishable based on the value of $\beta$:

- $\beta = -1$: scientific plausability is given zero attention and full weight lies on finding hypotheses near human reasoners, i.e., maximizing the likelihood that the algorithm's predictions will be discovered by human scientists in the near future.

- $\beta = 0$: scientific plausibility is given full emphasis and human availability of predictions is ignored.

- $\beta = 1$: scientific plausibility is ignored and the full weight rests on human avoidance, i.e., maximizing the likelihood that predictions will escape human scientists' collective attention.

In addition to these special conditions, we desire an algorithm where contribution from the two sources (plausibility and human accessibility) become equal when $|\beta| = \frac{1}{2}$, and the output score varies continuously as $\beta$ shifts.

For any given candidate material $x$, let us denote its SP-$d$ value and semantic similarity with respect to the property under consideration by $s_1(x)$ and $s_2(x)$, respectively. These scores have distinct units and vary at different scales, therefore a naive $\beta$-weighted averaging is inappropriate as it does not lead to equal contribution when $|\beta| = 1/2$. Moreover, the SP-$d$ values are unbounded as they can become arbitrarily large for entities disconnected from the property node in our hypergraph. As a result, Z-scores could not be directly applied. Instead, we applied a Van der Waerden transformation to first standardize the scores. Suppose $S$ is a set of scores and $s(x) \in S$, then the Van der Waerden transformation of $x$, denoted by $\tilde{s}(x)$, is defined as

$$\tilde{s}(x) \;=\; \phi\left(\frac{r(x)}{|S|+1}\right), \tag{S1}$$

where $\phi$ is the quantile function of the normal distribution, $r(x)$ is the rank of $s(x)$ within the set $S$ and $|S|$ denotes the cardinality of $S$. We then take the weighted average of Z-scores for the transformed signals $\tilde{s}_1(x)$ and $\tilde{s}_2(x)$ for each material $x$ as the ultimate hybrid score to be used in our final ranking. We further normalize the resulting transformed scores by computing their Z-scores yielding $\hat{s}_1(x)$ and $\hat{s}_2(x)$ before combining them in the following weighted averaging:

$$s_{\text{final}}(x) = \beta \hat{s}_1(x) + (1 - |\beta|)\hat{s}_2(x). \tag{S2}$$

Note that when $\beta < 0$, the algorithm tends to select materials with lower $\hat{s}_1(x)$, which in turn implies smaller SP-$d$ and materials with more contextual familiarity in terms of property. Alternatively, when $\beta > 0$, the algorithm scores higher those materials with a greater SP-$d$ and more unfamiliar or "alien" predictions will result.

## S2  Expectation Gap

The expectation gap is designed to show that while the percentage of predictions that become discovered after prediction year sharply decays with $\beta$, the quality of predictions remains high for substantially longer. We define two distributions over $\beta$ conditioned on (1) discoverability and (2) plausiability. Separation between the centers of these two distributions such that the latter is more right-skewed (decaying at higher values of $\beta$) functions as our indicator, defined as the gap between their expectations.

For a fixed property and pool of candidate materials, let us denote the set of all materials to be discovered after prediction year by $\mathcal{D}$ and the set of all plausible materials by $\mathcal{P}$. Also, let $\mathcal{H}_\beta$ and $h_\beta$ be the full set of predictions and a randomly selected prediction generated by our algorithm operating with $\beta$, respectively. Precision of the algorithm in terms of identifying near-future discoveries is defined as $\mathbb{P}(h_\beta \in \mathcal{D}|\beta)$, which can simply be computed by dividing the count of discoveries by the number of predictions: $\frac{|\mathcal{H}_\beta \cap \mathcal{D}|}{|\mathcal{H}_\beta|}$. Now using a uniform prior distribution over $\beta$, i.e., $\mathbb{P}(\beta) =$const., and applying Bayes rule, these precisions can be converted to $\mathbb{P}(\beta|h_\beta \in \mathcal{D})$ by normalization across all $\beta$ values such that they sum to unity.

Computing the second distribution is not as simple. The difficulty arises as we do not fully know $\mathcal{P}$, but rather have one real-valued score per material characterizing the likelihood of its $\mathcal{P}$-membership. These scores, denoted by $\tau$, are obtained from field-related theoretical knowledge and first-principles laws (see next section). In the first step, we transform theoretical scores to probabilities, such that for every material $x$, $\tau = \tau(x)$ goes to $\mathbb{P}(x \in \mathcal{P})$. Let $\tau_{\min}$ and $\tau_{\max}$ be the global minimum and maximum among all theoretical scores. We engineered a monotonically increasing transformation $T$ in the form of logit $\left[\tan\left(\pi\left(\hat{\tau} - \frac{1}{2}\right)\right) + b\right]$, where $\hat{\tau} = \frac{\tau - \tau_{\min}}{\tau_{\max} - \tau_{\min}}$ such that

- $T(\tau_{\min}) = 0$

- $T(\tau_{\max}) = 1$

- $T(\tau_{\text{mid}}) = \frac{1}{2}$, where $\tau_{\text{mid}} = \frac{1}{|\mathcal{D}|}\sum_{x \in \mathcal{D}} \tau(x)$, which is the average theoretical score attributed to discovered materials. This condition uniquely specifies parameter $b$.

Resulting probabilities will be thresholded by $\frac{1}{2}$ to probabilistically indicate which materials belong to $\mathbb{P}$. Setting the mid-point as above is a direct consequence of the assumption that the majority of materials discovered by scientists are plausible, hence we take their average theoretical scores as a baseline and every material with a higher $\tau$ will also be considered plausible. Such probabilistic classification of a material $x$ to $\mathcal{P}$ is done with a confidence level proportional to the distance between the probability $T(\tau(x))$ and the threshold $\frac{1}{2}$. The confidence level of our decision regarding $\mathcal{P}$-membership of sample $x$ is

$$c(x) = \begin{cases} T(\tau(x)) & , T(\tau(x)) \geq \frac{1}{2} \\ 1 - T(\tau(x)) & , T(\tau(x)) < \frac{1}{2} \end{cases} \tag{S3}$$

Now, for any prediction set $\mathcal{H}_\beta$, we use weighted maximum likelihood estimation to compute the probability of plausibility given $\beta$:

$$\mathbb{P}(h_\beta \in \mathcal{P}|\beta) = \frac{\sum_{x \in \mathcal{H}_\beta:\ T(\tau(x)) \geq 1/2} c(x)}{\sum_{x \in \mathcal{H}_\beta} c(x)}. \tag{S4}$$

Finally, similar to the previous case, the likelihood of $\beta$ given plausibility, $\mathbb{P}(\beta|h_\beta \in \mathcal{P})$, can be obtained by simply normalizing these probabilities across all $\beta$ values such that they sum to one.

We define the expectation gap as the difference between mean values of the two likelihoods described above:

$$\Delta\mathbb{E}[\beta] := \mathbb{E}[\beta|h_\beta \in \mathcal{P}] - \mathbb{E}[\beta|h_\beta \in \mathcal{D}]. \tag{S5}$$

Having a positive gap suggests that theoretical plausibility is higher for more alien predictions than those made and published by human scientists. Because discovery precision goes down with $\beta$, a positive expectation gap also means that there exists a non-empty gap when our algorithm approaches its ideal mission of human complementarity. Zero or negative gaps only occur for a few human diseases.

We evaluate expectation gaps from our algorithm for all properties with available theoretical scores $\tau$, i.e., thermoelectricity, ferroelectricity, COVID-19 and 45 other human disease (see section S4). We set the prediction year to 2001 for all but COVID-19 for which we ran our algorithm at the beginning of 2020. The results for prediction experiments are reported at the end of 2017 (for energy-related properties), the end of 2018 (for human diseases except COVID-19), and the end of July 2021 (for COVID-19). Note that here we did not perform prediction in the intermediate years between 2001 and 2017 for thermoelectricity.

## S3 Joint Probabilities

Our expectation gap provided us with a single evaluation score for the performance of our algorithm. However, it does not say anything about the desired range of $\beta$ where the algorithm operates closest to its mission to serve complementary, high-value predictions. Here, to provide a clearer overview of the performance of our algorithm for different $\beta$ values, we directly model and calculate the probability that it outputs unfamiliar yet scientifically promising (plausible) predictions. As described above, unfamiliarity of a random prediction $h_\beta$ means its unimaginability in context of the considered property and therefore its undiscoverability (i.e., $h_\beta \notin \mathcal{D}$). Thus, we calculate the probability that $h_\beta$ is unfamiliar and plausible given a certain $\beta$ by the joint distribution $\mathbb{P}(h_\beta \notin \mathcal{D}, h_\beta \in \mathcal{P}|\beta)$. Applying Bayes rule, this joint probability decomposes into two simpler distributions:

$$\mathbb{P}(h_\beta \notin \mathcal{D}, h_\beta \in \mathcal{P}|\beta) = \mathbb{P}(h_\beta \notin \mathcal{D}|\beta) \cdot \mathbb{P}(h_\beta \in \mathcal{P}|\beta, h_\beta \notin \mathcal{D}), \tag{S6}$$

where the first term in the right-hand-side is the complementary of discovery precision $(1 - \mathbb{P}(h_\beta \in \mathcal{D}|\beta))$ and the second term can be computed similarly to the probability of plausibility, $\mathbb{P}(h_\beta \in \mathcal{P}|\beta)$, described above. Specifically, all computations are to be repeated on those predictions *not* discovered after prediction year ($h_\beta \notin \mathcal{D}$), hence replacing $\mathcal{H}_\beta$ with $\mathcal{H}_\beta - \mathcal{D}$ in equation (S4).

# S4 Theoretical Scoring of Candidate Hypotheses

In order to assess the quality of predictions resulting from our algorithm, we used theoretically driven scores derived from first-principles equations or simulation models from the relevant disciplines. The underlying procedure for defining and curating such data naturally differs for disinct properties. In this section, we describe these theoretical scores for properties about which datasets could be found.

## S4.1 Thermoelectricity

Thermoelectricity is the property of producing electrical voltage when temperature varies on two sides of a material. An important measure of thermoelectricity that depends on the Seebeck coefficient and electrical conductivity is called Power Factor (PF). PF is also the major part of another common dimensionless criterion named its *figure of merit (zT)*. First-principles methods based on Density Functional Theory (DFT) have been widely used to estimate energy-related chemical properties including PF[2]. Recently, DFT-based PF estimates have been used to evaluate content-based discovery predictions[4], where it is shown that materials studied in connection with thermoelectricity have larger PF values on average than unstudied materials. Here, we use the same set of PF estimates, which have been prepared by taking the maximum of the average PFs computed for various temperatures, doping levels and semiconductor types.

When running our algorithm at prediction year 2001, we only considered those materials mentioned at least once within the five-year period [1996, 2000] preceding 2001 (12.6K materials) and are assigned a PF estimate (13.6K materials). Applying both conditions at the same time reduced the size of our pool of materials to approximately 3.5K.

## S4.2 Ferroelectricity

Ferroelectric materials are characterized by their spontaneous electric polarization, which is reversible in the presence of an external electric field. Recently, Smidt et al. developed an automated workflow that uses symmetry analysis and first-principles calculations to curate a list of ferroelectric materials[3]. Their final list included 255 candidates, where entries are rated based on the magnitude of their spontaneous polarization. There are 167 distinct chemical formulae in this list, and for some several ferroelectricity scores are reported corresponding to various crystal structures. For compounds with multiple values, we considered the maximum computed ferroelectricity over all available structures as the final score.

When running our algorithm with ferroelectricity, we set the prediction year to 2001 and considered 2,551 materials that appeared in at least one paper in the five-year period [1996, 2000]. Only 167 distinct compounds that have been reported as ferroelectric by Smidt et al. are assigned non-zero scores as explained above and the rest are considered to have zero ferroelectricity. This leads to a sparse scoring system but, as is shown in our results, remains nevertheless sufficient to demonstrate the performance of our method.

## S4.3  Human Diseases

The similarity between two sets of proteins targeted by a particular disease and a certain drug forms a basis for measuring their underlying association and therapeutic potential. Protein-protein interaction networks include protein, drug and disease nodes with pairwise interactions encoded within the edges. We utilize drug-disease proximity in such networks as the core of our theoretical scoring framework to assess drug candidates in terms of their relevance to the treatment of a disease. Recently, Gysi et al. showed that drugs whose target proteins are within or in vicinity of the COVID-19 disease module are potentially strong candidates for repurposing to treat or prevent the infection[1]. They used various proximity measures to compute network similarities and identify the most relevant drug candidates. Among these measures was cosine similarity of embedding vectors resulting from a pretrained Graph Neural Network (GNN) over the protein-protein network, which we used in our evaluations of drug-disease proposals.

The GNN-based embedding vectors of the interaction network produced and shared by Gysi et al.[1] included 1.6K drug nodes and 2.5K disease nodes (including COVID-19), among which 1.5K drugs belonged to our pool of candidates and 45 diseases were common with the set of 100 human diseases we considered in our experiments. We ran our complementary algorithm at prediction year 2001 for all diseases but COVID-19, where we restricted our pool of candidates to 1,179 drugs that existed in the protein-protein network and also appeared in the literature between 1996 and 2000. For COVID-19, the prediction year is set to the beginning of 2020 with 1,436 candidate drugs that existed in both the interaction network and in the literature between 2015 and 2019.

# S5  Complementary Prediction on the Full Set of Human Diseases

In this section, we display the result of running our complementary discovery prediction on the full set of human diseases (176 cases with available protein-protein theoretical scores). Table 1 shows overlapping percentage between our algorithm's predicted and the actual discoveries together with the average theoretical protein-protein similarity scores for candidates generated with $\beta$ values 0, 0.2 and 0.4. An ideal set of complementary discovery candidates is expected to have low overlapping percentage (low cognitive availability) and high theoretical scores (scientific merit).

Table 1: Evaluating predicted discoveries by our complementary discovery prediction algorithms operated with $\beta$ values 0, 0.2 and 0.4.

| Disease | $\beta = 0$ | | $\beta = 0.2$ | | $\beta = 0.4$ | |
|---|---|---|---|---|---|---|
| | Overlap (%) | Pr-Pr Sim. | Overlap (%) | Pr-Pr Sim. | Overlap (%) | Pr-Pr Sim. |
| Acromegaly | 0 | 0.323 | 0 | 0.320 | 0 | 0.259 |
| Adenocarcinoma | 11 | 0.066 | 10 | 0.009 | 13 | -0.133 |
| Alcoholism | 23 | 0.565 | 19 | 0.539 | 15 | 0.475 |
| Alopecia | 37 | 0.152 | 26 | 0.165 | 25 | 0.200 |
| Amenorrhea | 3 | 0.264 | 3 | 0.268 | 1 | 0.273 |
| Amyloidosis | 22 | 0.269 | 18 | 0.257 | 16 | 0.253 |
| Anaphylaxis | 4 | 0.367 | 3 | 0.369 | 1 | 0.335 |
| Anemia | 22 | 0.155 | 21 | 0.150 | 22 | 0.168 |
| Angioedema | 11 | 0.183 | 10 | 0.205 | 5 | 0.225 |
| Appendicitis | 19 | 0.281 | 13 | 0.222 | 8 | 0.145 |
| Arteriosclerosis | 41 | 0.441 | 35 | 0.423 | 28 | 0.406 |
| Arthritis | 13 | 0.309 | 9 | 0.276 | 6 | 0.129 |
| Arthrogryposis | 2 | 0.363 | 2 | 0.364 | 1 | 0.305 |
| Ascites | 6 | 0.142 | 5 | 0.176 | 1 | 0.245 |
| Asthma | 10 | 0.351 | 9 | 0.329 | 16 | 0.243 |
| Astrocytoma | 39 | 0.169 | 32 | 0.165 | 25 | 0.111 |
| Ataxia | 23 | 0.423 | 17 | 0.405 | 12 | 0.333 |
| Atherosclerosis | 33 | 0.387 | 35 | 0.400 | 32 | 0.281 |
| Blepharospasm | 2 | 0.305 | 1 | 0.259 | 1 | 0.140 |
| Blindness | 0 | 0.219 | 1 | 0.212 | 1 | 0.208 |
| Brachydactyly | 4 | 0.238 | 4 | 0.164 | 1 | 0.036 |
| Bronchiectasis | 36 | 0.408 | 34 | 0.387 | 20 | 0.341 |

Table 1 (continued)

| Disease | $\beta = 0$ | | $\beta = 0.2$ | | $\beta = 0.4$ | |
|---|---|---|---|---|---|---|
| | Compl. (%) | Pr-Pr Sim. | Compl. (%) | Pr-Pr Sim. | Compl. (%) | Pr-Pr Sim. |
| Brucellosis | 2 | 0.288 | 2 | 0.215 | 1 | 0.065 |
| Candidiasis | 6 | 0.225 | 5 | 0.155 | 0 | 0.069 |
| Carcinoma | 15 | 0.163 | 12 | 0.086 | 10 | -0.176 |
| Cardiomyopathies | 15 | 0.393 | 15 | 0.379 | 11 | 0.383 |
| Cataract | 15 | 0.327 | 12 | 0.327 | 9 | 0.314 |
| Cholangiocarcinoma | 48 | 0.213 | 42 | 0.168 | 30 | 0.001 |
| Cholangitis | 10 | 0.217 | 8 | 0.164 | 6 | 0.097 |
| Cholelithiasis | 6 | 0.220 | 6 | 0.171 | 5 | 0.106 |
| Cholestasis | 23 | 0.255 | 26 | 0.241 | 32 | 0.239 |
| Chorioamnionitis | 17 | 0.274 | 13 | 0.236 | 11 | 0.194 |
| Colitis | 41 | 0.256 | 36 | 0.262 | 31 | 0.223 |
| Constipation | 19 | 0.334 | 16 | 0.329 | 10 | 0.304 |
| Contracture | 0 | 0.408 | 0 | 0.362 | 0 | 0.329 |
| Cryptorchidism | 43 | 0.350 | 35 | 0.337 | 17 | 0.260 |
| Cystitis | 11 | 0.361 | 12 | 0.362 | 9 | 0.283 |
| Delirium | 10 | 0.378 | 10 | 0.360 | 8 | 0.296 |
| Dermatitis | 17 | 0.311 | 13 | 0.252 | 8 | 0.206 |
| Dermatomyositis | 37 | 0.216 | 30 | 0.224 | 21 | 0.214 |
| Diarrhea | 8 | 0.267 | 10 | 0.283 | 8 | 0.251 |
| Dizziness | 6 | 0.551 | 6 | 0.535 | 3 | 0.433 |
| Dysarthria | 6 | 0.359 | 4 | 0.315 | 2 | 0.277 |
| Dyskinesias | 35 | 0.623 | 34 | 0.610 | 25 | 0.518 |
| Dyslipidemias | 35 | 0.377 | 28 | 0.356 | 14 | 0.260 |
| Dyspnea | 13 | 0.400 | 10 | 0.430 | 9 | 0.425 |
| Eczema | 7 | 0.225 | 6 | 0.215 | 6 | 0.126 |
| Embolism | 3 | 0.246 | 2 | 0.226 | 2 | 0.202 |

Table 1 (continued)

| Disease | $\beta = 0$ | | $\beta = 0.2$ | | $\beta = 0.4$ | |
|---|---|---|---|---|---|---|
| | Compl. (%) | Pr-Pr Sim. | Compl. (%) | Pr-Pr Sim. | Compl. (%) | Pr-Pr Sim. |
| Emphysema | 22 | 0.378 | 21 | 0.330 | 17 | 0.312 |
| Endometriosis | 55 | 0.327 | 46 | 0.314 | 39 | 0.195 |
| Entamoebiasis | 31 | 0.022 | 24 | -0.050 | 12 | -0.110 |
| Enterocolitis | 12 | 0.092 | 13 | 0.100 | 9 | 0.107 |
| Eosinophilia | 9 | 0.240 | 7 | 0.215 | 6 | 0.136 |
| Epilepsy | 17 | 0.457 | 14 | 0.436 | 15 | 0.331 |
| Exanthema | 11 | 0.108 | 10 | 0.092 | 9 | 0.083 |
| Gallstones | 4 | 0.216 | 2 | 0.181 | 1 | 0.101 |
| Gastroenteritis | 3 | -0.081 | 2 | -0.069 | 1 | -0.006 |
| Gastroparesis | 15 | 0.180 | 12 | 0.180 | 9 | 0.186 |
| Glaucoma | 14 | 0.358 | 10 | 0.383 | 8 | 0.339 |
| Glioblastoma | 33 | 0.336 | 25 | 0.301 | 13 | 0.143 |
| Glioma | 35 | 0.223 | 30 | 0.152 | 28 | -0.002 |
| Gliosarcoma | 20 | 0.180 | 18 | 0.148 | 12 | 0.122 |
| Glomerulonephritis | 9 | 0.226 | 7 | 0.206 | 5 | 0.176 |
| Goiter | 3 | 0.009 | 2 | -0.011 | 2 | -0.002 |
| Gout | 39 | 0.327 | 35 | 0.286 | 27 | 0.264 |
| Heartburn | 3 | 0.159 | 3 | 0.147 | 2 | 0.142 |
| Hemangioblastoma | 17 | -0.079 | 13 | -0.100 | 5 | -0.135 |
| Hemangioma | 5 | 0.172 | 5 | 0.174 | 3 | 0.131 |
| Hemorrhage | 4 | 0.263 | 4 | 0.263 | 3 | 0.233 |
| Hemorrhoids | 3 | 0.198 | 3 | 0.188 | 2 | 0.205 |
| Hyperaldosteronism | 6 | 0.429 | 4 | 0.418 | 2 | 0.398 |
| Hyperalgesia | 21 | 0.522 | 20 | 0.510 | 19 | 0.432 |
| Hyperammonemia | 4 | 0.206 | 4 | 0.210 | 3 | 0.205 |
| Hypercholesterolemia | 25 | 0.383 | 24 | 0.377 | 19 | 0.295 |

Table 1 (continued)

| Disease | $\beta = 0$ | | $\beta = 0.2$ | | $\beta = 0.4$ | |
|---|---|---|---|---|---|---|
| | Compl. (%) | Pr-Pr Sim. | Compl. (%) | Pr-Pr Sim. | Compl. (%) | Pr-Pr Sim. |
| Hyperinsulinism | 10 | 0.280 | 10 | 0.271 | 7 | 0.245 |
| Hyperkalemia | 8 | 0.182 | 5 | 0.171 | 5 | 0.183 |
| Hyperlipidemias | 32 | 0.423 | 27 | 0.397 | 13 | 0.277 |
| Hyperparathyroidism | 13 | 0.243 | 12 | 0.245 | 11 | 0.191 |
| Hyperphosphatemia | 7 | 0.413 | 6 | 0.408 | 4 | 0.376 |
| Hyperpigmentation | 11 | 0.234 | 10 | 0.224 | 7 | 0.198 |
| Hyperplasia | 34 | 0.165 | 35 | 0.172 | 31 | 0.198 |
| Hyperprolactinemia | 2 | 0.206 | 1 | 0.175 | 0 | 0.110 |
| Hypersensitivity | 19 | 0.230 | 20 | 0.220 | 24 | 0.189 |
| Hypertension | 3 | 0.553 | 5 | 0.490 | 9 | 0.361 |
| Hyperthyroidism | 18 | 0.040 | 16 | 0.039 | 16 | 0.027 |
| Hypertrophy | 41 | 0.394 | 38 | 0.394 | 25 | 0.320 |
| Hyperuricemia | 22 | 0.317 | 17 | 0.350 | 11 | 0.262 |
| Hypoalbuminemia | 19 | 0.185 | 18 | 0.175 | 14 | 0.124 |
| Hypocalcemia | 5 | 0.232 | 3 | 0.242 | 2 | 0.227 |
| Hypoglycemia | 4 | 0.228 | 3 | 0.205 | 3 | 0.255 |
| Hypogonadism | 5 | 0.213 | 8 | 0.169 | 6 | 0.095 |
| Hypokalemia | 4 | 0.396 | 3 | 0.387 | 2 | 0.339 |
| Hyponatremia | 7 | 0.289 | 6 | 0.264 | 6 | 0.241 |
| Hypoparathyroidism | 2 | 0.217 | 2 | 0.243 | 1 | 0.177 |
| Hypophosphatasia | 16 | -0.133 | 13 | -0.124 | 7 | -0.102 |
| Hypophosphatemia | 8 | 0.254 | 8 | 0.230 | 5 | 0.172 |
| Hypopigmentation | 3 | 0.091 | 3 | 0.103 | 2 | 0.175 |
| Hypopituitarism | 16 | 0.218 | 15 | 0.206 | 13 | 0.196 |
| Hypotension | 6 | 0.520 | 6 | 0.495 | 8 | 0.423 |
| Hypothyroidism | 7 | 0.235 | 3 | 0.208 | 2 | 0.188 |

Table 1 (continued)

| Disease | $\beta = 0$ | | $\beta = 0.2$ | | $\beta = 0.4$ | |
|---|---|---|---|---|---|---|
| | Compl. (%) | Pr-Pr Sim. | Compl. (%) | Pr-Pr Sim. | Compl. (%) | Pr-Pr Sim. |
| Hypotrichosis | 3 | 0.284 | 1 | 0.263 | 0 | 0.207 |
| Hypoxia | 12 | 0.361 | 10 | 0.373 | 8 | 0.332 |
| Keloid | 37 | 0.291 | 34 | 0.268 | 20 | 0.175 |
| Keratitis | 22 | 0.270 | 13 | 0.193 | 7 | 0.138 |
| Keratosis | 38 | 0.152 | 31 | 0.131 | 15 | 0.134 |
| Leiomyoma | 21 | 0.141 | 17 | 0.104 | 12 | -0.002 |
| Leiomyosarcoma | 30 | 0.219 | 28 | 0.126 | 18 | -0.008 |
| Leishmaniasis | 52 | 0.305 | 43 | 0.249 | 28 | 0.165 |
| Leprosy | 3 | 0.089 | 2 | 0.065 | 1 | 0.021 |
| Leukemia | 21 | 0.173 | 16 | 0.111 | 9 | 0.023 |
| Liposarcoma | 14 | 0.145 | 12 | 0.079 | 4 | -0.119 |
| Listeriosis | 30 | 0.274 | 26 | 0.219 | 17 | 0.012 |
| Lymphangioleiomyomatosis | 10 | 0.028 | 8 | -0.014 | 2 | -0.144 |
| Lymphoma | 16 | 0.290 | 17 | 0.224 | 15 | 0.064 |
| Malaria | 7 | 0.241 | 4 | 0.220 | 0 | 0.129 |
| Malnutrition | 8 | 0.078 | 6 | 0.066 | 4 | 0.050 |
| Medulloblastoma | 35 | 0.291 | 28 | 0.224 | 20 | 0.093 |
| Melanoma | 24 | 0.257 | 19 | 0.232 | 15 | 0.145 |
| Meningitis | 3 | 0.169 | 3 | 0.105 | 1 | 0.042 |
| Methemoglobinemia | 16 | 0.196 | 16 | 0.184 | 12 | 0.116 |
| Mucositis | 17 | 0.159 | 18 | 0.125 | 15 | 0.038 |
| Myoclonus | 15 | 0.647 | 12 | 0.627 | 10 | 0.515 |
| Myositis | 2 | 0.261 | 3 | 0.223 | 1 | 0.247 |
| Narcolepsy | 11 | 0.504 | 10 | 0.486 | 5 | 0.352 |
| Neoplasms | 23 | 0.200 | 26 | 0.090 | 24 | 0.006 |
| Nephritis | 9 | 0.219 | 11 | 0.226 | 4 | 0.178 |

Table 1 (continued)

| Disease | β = 0 | | β = 0.2 | | β = 0.4 | |
|---|---|---|---|---|---|---|
| | Compl. (%) | Pr-Pr Sim. | Compl. (%) | Pr-Pr Sim. | Compl. (%) | Pr-Pr Sim. |
| Nephrolithiasis | 4 | 0.415 | 6 | 0.400 | 2 | 0.332 |
| Nephrosis | 6 | 0.328 | 5 | 0.339 | 5 | 0.273 |
| Neuroblastoma | 13 | 0.253 | 11 | 0.267 | 10 | 0.096 |
| Neurofibrosarcoma | 3 | 0.151 | 2 | 0.110 | 2 | -0.035 |
| Neutropenia | 11 | 0.228 | 9 | 0.171 | 8 | 0.089 |
| Ophthalmoplegia | 3 | 0.147 | 3 | 0.166 | 3 | 0.134 |
| Osteoarthritis | 30 | 0.342 | 27 | 0.311 | 27 | 0.152 |
| Osteoporosis | 35 | 0.398 | 33 | 0.372 | 26 | 0.291 |
| Osteosarcoma | 56 | 0.349 | 45 | 0.290 | 36 | 0.163 |
| Pancreatitis | 41 | 0.175 | 40 | 0.159 | 33 | 0.166 |
| Paresis | 1 | 0.339 | 0 | 0.305 | 0 | 0.273 |
| Periodontitis | 17 | 0.315 | 11 | 0.303 | 8 | 0.259 |
| Peritonitis | 16 | 0.346 | 15 | 0.324 | 12 | 0.200 |
| Pheochromocytoma | 14 | 0.278 | 13 | 0.275 | 9 | 0.260 |
| Photophobia | 4 | 0.244 | 2 | 0.225 | 2 | 0.204 |
| Pneumonia | 32 | 0.424 | 32 | 0.363 | 22 | 0.211 |
| Polyneuropathies | 10 | 0.360 | 8 | 0.346 | 3 | 0.303 |
| Polyuria | 5 | 0.396 | 3 | 0.387 | 2 | 0.351 |
| Pre-Eclampsia | 19 | 0.362 | 17 | 0.348 | 15 | 0.341 |
| Prolactinoma | 2 | 0.234 | 1 | 0.208 | 0 | 0.168 |
| Prostatitis | 10 | 0.477 | 9 | 0.421 | 4 | 0.285 |
| Proteinuria | 8 | -0.022 | 10 | 0.016 | 10 | 0.088 |
| Pruritus | 30 | 0.294 | 31 | 0.290 | 22 | 0.238 |
| Psoriasis | 5 | 0.324 | 7 | 0.287 | 10 | 0.165 |
| Pyelonephritis | 1 | 0.472 | 1 | 0.366 | 1 | 0.173 |
| Rhabdomyosarcoma | 18 | 0.292 | 14 | 0.218 | 9 | 0.066 |

Table 1 (continued)

| Disease | $\beta = 0$ | | $\beta = 0.2$ | | $\beta = 0.4$ | |
|---|---|---|---|---|---|---|
| | Compl. (%) | Pr-Pr Sim. | Compl. (%) | Pr-Pr Sim. | Compl. (%) | Pr-Pr Sim. |
| Rhinitis | 30 | 0.473 | 20 | 0.421 | 8 | 0.293 |
| Sarcoidosis | 10 | 0.200 | 8 | 0.194 | 7 | 0.120 |
| Sarcoma | 16 | 0.314 | 11 | 0.265 | 8 | 0.076 |
| Schistosomiasis | 6 | 0.153 | 6 | 0.147 | 5 | 0.105 |
| Schizophrenia | 18 | 0.659 | 18 | 0.573 | 23 | 0.463 |
| Seizures | 6 | 0.608 | 10 | 0.603 | 15 | 0.482 |
| Seminoma | 13 | 0.153 | 8 | 0.107 | 5 | -0.101 |
| Sinusitis | 5 | 0.460 | 4 | 0.379 | 2 | 0.272 |
| Stomatitis | 9 | 0.064 | 7 | 0.095 | 2 | 0.079 |
| Stroke | 17 | 0.399 | 15 | 0.401 | 9 | 0.280 |
| Synovitis | 10 | 0.234 | 7 | 0.218 | 4 | 0.150 |
| Tetany | 3 | 0.162 | 3 | 0.165 | 2 | 0.126 |
| Thrombocytopenia | 14 | 0.252 | 11 | 0.256 | 5 | 0.234 |
| Thromboembolism | 8 | 0.215 | 8 | 0.183 | 5 | 0.141 |
| Thrombosis | 9 | 0.294 | 18 | 0.296 | 18 | 0.255 |
| Toothache | 2 | 0.170 | 2 | 0.155 | 0 | 0.046 |
| Torticollis | 4 | 0.462 | 3 | 0.409 | 1 | 0.263 |
| Tremor | 11 | 0.600 | 9 | 0.590 | 8 | 0.534 |
| Trichuriasis | 18 | 0.161 | 15 | 0.157 | 9 | 0.184 |
| Tuberculosis | 5 | 0.269 | 4 | 0.243 | 3 | 0.193 |
| Urticaria | 26 | 0.329 | 23 | 0.333 | 18 | 0.282 |
| Uveitis | 13 | 0.218 | 12 | 0.195 | 7 | 0.192 |
| Vitiligo | 46 | 0.167 | 39 | 0.150 | 25 | 0.095 |
| Xerostomia | 4 | 0.220 | 4 | 0.214 | 3 | 0.139 |

# References

[1] D. M. Gysi, Í. Do Valle, M. Zitnik, A. Ameli, X. Gan, O. Varol, S. D. Ghiassian, J. Patten, R. A. Davey, J. Loscalzo, et al. Network medicine framework for identifying drug-repurposing opportunities for covid-19. *Proceedings of the National Academy of Sciences*, 118(19), 2021.

[2] F. Ricci, W. Chen, U. Aydemir, G. J. Snyder, G.-M. Rignanese, A. Jain, and G. Hautier. An ab initio electronic transport database for inorganic materials. *Scientific data*, 4(1): 1–13, 2017.

[3] T. E. Smidt, S. A. Mack, S. E. Reyes-Lillo, A. Jain, and J. B. Neaton. An automatically curated first-principles database of ferroelectrics. *Scientific data*, 7(1):1–22, 2020.

[4] V. Tshitoyan, J. Dagdelen, L. Weston, A. Dunn, Z. Rong, O. Kononova, K. A. Persson, G. Ceder, and A. Jain. Unsupervised word embeddings capture latent knowledge from materials science literature. *Nature*, 571(7763):95–98, 2019.